\documentclass[conference]{IEEEtran}
\IEEEoverridecommandlockouts
\usepackage{cite}
\usepackage{amsmath,amssymb,amsfonts}
\usepackage{algorithmic}
\usepackage{graphicx}
\usepackage{textcomp}
\usepackage{xcolor}
\def\BibTeX{{\rm B\kern-.05em{\sc i\kern-.025em b}\kern-.08em
    T\kern-.1667em\lower.7ex\hbox{E}\kern-.125emX}}

\newcommand{\etal}{\textit{et al.}}
\newtheorem{definition}{Definition}
\usepackage{color} 
\usepackage{tabularx} 
\usepackage{adjustbox} 
\usepackage{multirow} 
\usepackage{subcaption}
\usepackage{hyperref}

\begin{document}

\title{Towards Real-Time Temporal Graph Learning}
\author{
  \IEEEauthorblockN{
    Deniz Gurevin\IEEEauthorrefmark{1},
    Mohsin Shan\IEEEauthorrefmark{1},
    Tong Geng\IEEEauthorrefmark{2},
    Weiwen Jiang\IEEEauthorrefmark{3},
    Caiwen Ding\IEEEauthorrefmark{1} and
    Omer Khan\IEEEauthorrefmark{1}
  }

   \IEEEauthorblockA{\IEEEauthorrefmark{1}University of Connecticut, Storrs, CT, USA }
   \IEEEauthorblockA{\IEEEauthorrefmark{2}University of Rochester,  Rochester, NY, USA }
   \IEEEauthorblockA{\IEEEauthorrefmark{3}George Mason University, Fairfax, VA,  USA \\
       \IEEEauthorrefmark{1}\{deniz.gurevin, mohsin.shan, caiwen.ding, khan\}@uconn.edu, 
    \IEEEauthorrefmark{2}tgeng@ur.rochester.edu, 
    \IEEEauthorrefmark{3}wjiang8@gmu.edu
    \vspace{-15pt}
    }

}
\maketitle

\begin{abstract}

In recent years, graph representation learning has gained significant popularity, which aims to generate node embeddings that capture features of graphs. One of the methods to achieve this is employing a technique called \textit{random walks} that captures node sequences in a graph and then learns embeddings for each node using a natural language processing technique called Word2Vec. These embeddings are then used for deep learning on graph data for classification tasks, such as link prediction or node classification. Prior work operates on pre-collected temporal graph data and is not designed to handle updates on a graph in real-time. Real world graphs change dynamically and their entire temporal updates are not available upfront. In this paper, we propose an end-to-end graph learning pipeline that performs temporal graph construction, creates low-dimensional node embeddings, and trains multi-layer neural network models in an online setting. The training of the neural network models is identified as the main performance bottleneck as it performs repeated matrix operations on many sequentially connected low-dimensional kernels. We propose to unlock fine-grain parallelism in these low-dimensional kernels to boost performance of model training.

\end{abstract}


\section{Introduction}

Graph representation learning (GRL) utilizes artificial intelligence methods to learn the representation of graph structured data \cite{Scarselli2009TheGN, Hamilton2017InductiveRL, Velickovic2018GraphAN, Perozzi2014DeepWalkOL, Ahmed2019role2vecRN, Ribeiro2017struc2vecLN}. It has gained significant popularity in various application areas from social networks\cite{Backstrom2011SupervisedRW, Bian2020RumorDO}, to biology and chemistry\cite{Choma2018GraphNN,Duvenaud2015ConvolutionalNO,Stokes2020ADL}. Although prior works have analyzed the performance of GRL workloads \cite{Yan2020CharacterizingAU, Baruah2021GNNMarkAB,Zhang2020ArchitecturalIO}, they have mostly focused on static input graphs using graph representation learning techniques such as Graph Convolution Network (GCN) \cite{Kipf2017SemiSupervisedCW} and others\cite{Hamilton2017InductiveRL, Xu2019HowPA}.

To learn graph dynamics on temporally changing graphs, random walk based GRL algorithms have been proposed \cite{Talati2021ADD}. Unlike static graphs, temporal graphs are dynamically changing graphs with time data associated with each interaction between nodes. The front-end of the pipeline takes a temporal graph as an input and maps the underlying graph structure into a low-dimension embedding space by feeding temporal random walks  \cite{Nguyen2018ContinuousTimeDN} into a word2vec model, which is a common technique from Natural Language Processing (NLP) \cite{mikolov2013efficient}. This GRL process outputs \textit{node embeddings} that capture the underlying features of the nodes in the graph. These node embeddings are then used to train a Feed-forward Neural Network (FNN) for link prediction or node classification tasks.

A shortcoming of prior work \cite{Talati2021ADD} is that the input graph to the pipeline is not temporally updated. The workload takes graph data that contains the entire temporal information and performs the GRL algorithm to train an FNN on the pre-collected graph data. Therefore, the pipeline is not designed to assign temporal updates on the GRL algorithm and train FNNs on the fly with streaming snapshots of dynamic graphs. However, real world graphs change dynamically and their temporal updates are not available all at once, but sequentially collected between graph snapshots associated with timestamps \cite{Sahoo2018OnlineDL, Yoon2018LifelongLW, Lee2016DualMemoryDL}. At each timestamp, GRL needs to be performed only on the updated portion of the graph. More importantly, node embeddings that capture the entire history of graph dynamics are not available to the FNN prior to the training in a real world setting. As the graph evolves, embeddings also change temporally, and the FNN training at timestamp $t$ is performed with the node embeddings that contain information about only the current and previous timestamps \{$0$, $1$, ..., $t$\} and not future timestamps \{$t+1$, $t+2$, ..., $T$\}. Due to these reasons, it is computationally infeasible to perform the FNN training on the entire temporal graph data once. The training must be performed on discrete data batches that are collected in sequential order. In the previous work, this type of training approach has not been evaluated.

The main objective of this paper is to implement and characterize a temporally updating graph learning pipeline and perform the FNN training using an online setting where the training data becomes available as the graph evolves. Our pipeline takes temporal graph snapshots at each timestamp $t$, and starts with an R-Tree based \textit{graph construction step} \cite{Gurevin2021AnEA} that can keep track of the new updates in the graph since the last timestamp $t-1$. These updates are then streamed into the \textit{GRL step} that performs random walks and word2vec on the graph to obtain node embeddings. We utilize EvoNRL\cite{Heidari2020EvolvingNR} that continuously learns embeddings from the temporally updated graph into low-dimension graph representations, while avoiding redundant updates between graph snapshots. The graph updates are then forwarded to the final \textit{training step} for link prediction or node classification. Here, we adopt an online learning strategy, where a single temporal graph batch is trained for consecutive iterations at each timestamp $t$. This step concludes the temporal graph learning pipeline for a timestamp. These steps are repeated for the subsequent timestamps, $\{t+1, t+2, ..., N\}$.

Our evaluation of the temporal GRL pipeline shows that the execution time is dominated by the FNN training phase. Although several parallelization methods are proposed for random walks and word2vec in \cite{Talati2021ADD}, the FNN training implementation is left out for future performance enhancements. In this paper, we explore fine-grain parallelization of FNN training for performance acceleration. Our strategy considers parallelization opportunities in the individual matrix kernels in the forward and backward propagation paths for each iteration of the FNN model. We implement four state-of-the-art matrix multiplication (MM) algorithms (i.e., inner, outer, row-wise and column-wise product) for each low-dimensional kernel in the FNN model. The evaluation on a large core count shared memory multicore shows that using the right parallelization strategy yields significant potential for performance scaling. In summary, we make the following contributions:
\begin{itemize}
    \item We propose an end-to-end temporal random walk-based graph learning algorithm for online processing of temporal graphs. 
    The proposed real-time GRL algorithm is implemented in Python, and released as an open-source benchmark at \url{https://github.com/grvndnz/ST-Graph-Learning}.
    The performance analysis identifies the FNN training as the main performance bottleneck. 
    \item We implement a C++ parallel implementation of the N-layer Feed-forward Neural Network (FNN) pipeline. 
    The individual matrix kernels in the forward and backward propagation steps of FNN training are evaluated for in-depth performance analysis.
    We implement four state-of-the-art parallelization strategies for the low-dimensional matrix-multiplications in the FNN pipeline to evaluate the performance scaling potential on multicore processors.
    The FNN pipeline implementations are released as an open-source benchmark at \url{https://github.com/grvndnz/Parallel-FNN}.
\end{itemize}

\section{Related Work} 
In the recent years, there has been a surge in research in the area of graph representation learning (GRL), which aims to encode graph structure into a low-dimensional embedding space \cite{Hamilton2017RepresentationLO}. The main goal of GRL research is to optimize this encoding in a way that the representations in the learned space reflect the original graph structure. Some of the early works in GRL include DeepWalk\cite{Perozzi2014DeepWalkOL} and Node2Vec\cite{Grover2016node2vecSF}, which leverage node proximity in graphs using the the idea of word proximity NLP \cite{mikolov2013efficient}. Later, other works have incorporated this idea to learn graph structural
properties such as similarity in degree sequences \cite{Ribeiro2017struc2vecLN} and the behavioral roles of the nodes \cite{Ahmed2019role2vecRN}. These embedding algorithms train node embeddings for individual nodes, and therefore, require additional training via stochastic gradient descent to  make predictions on new nodes. There are works for learning inductive node embeddings that combine external node features into graph structures. These works include graph convolutional networks (GCN) \cite{Kipf2017SemiSupervisedCW}, GNNs \cite{Scarselli2009TheGN}, GraphSAGE \cite{Hamilton2017InductiveRL}, and Graph Attention Networks (GAT) \cite{Velickovic2018GraphAN}.

Most research for deep learning on graphs assumes that the underlying graph is static. However, the idea of temporally changing graphs is more realistic when it comes to most real-life systems. While static graph learning models can be applied to temporal graphs by ignoring the temporal evolution \cite{LibenNowell2007TheLP}, temporal graph structure contains significant
insights about the system. There have been works that explore processing temporal graphs as a sequence of snapshots \cite{Sankar2020DySATDN, Goyal2020dyngraph2vecCN, Pareja2020EvolveGCNEG, Dunlavy2011TemporalLP} that can capture the evolution of temporal graph dynamics. Similarly, streaming graphs process temporally changing data in the finest granularity in terms of time and it is computationally much more expensive compared to snapshots \cite{Aggarwal2010OnCG}. 

Previously, in terms of processing temporal graphs for graph representation learning, Talati \etal \cite{Talati2021ADD} proposed an implementation using a temporal random-walk based algorithm. This work has performed detailed algorithm and hardware based performance characterization of the pipeline and identified several execution bottlenecks. However, the analyzed GRL and FNN training algorithms are performed on pre-collected graph data and not temporally evolving snapshots of graph. Consequently, the FNN model is trained with node embeddings that capture the entire history of the graph structure (past and future node interactions). In this work, we propose an implementation that evaluates prediction tasks in an online setting on temporally changing graphs.

\section{Run-time Temporal Graph Learning}\label{sec:proposed_methods}

In this section, we give the details of the proposed run-time GRL algorithm for temporal graphs. We begin with giving the following definitions.

\noindent \begin{definition}
A graph $G$ is defined as a tuple $G = (V, E)$ where $V = \{v_0, v_1, ..., v_n\}$ is the set of $n$ nodes and $E = \{e_0, e_1, ...,e_m\}$ is the set of $m$ edges where an edge connects two nodes.
\end{definition}

In the case of temporal graphs, where nodes are continuously added/removed and edges dynamically change with time, in order to maintain these temporal updates, the graph is processed into a sequence of discrete graph \textit{snapshots}. Therefore, a temporal network is traditionally represented as a sequence of static graphs $(G_1, G_2, ..., G_T)$ for $T$ timestamps. Similarly, an edge between node $v_a$ and node $v_b$ at a timestamp $t \in \{0, 1, ..., T-1\}$ is represented as $(v_a, v_b, t)$.

\noindent\begin{definition}
In a graph $G = (V, E)$, a random walk from node $v_a$ to node $v_b$ is defined as a sequence of connected edges $w = \{(v_a, v_1), (v_1, v_2), ..., (v_k,  v_b)\}$ where $k+1$ is the length of $w$.
\end{definition} 

The main concept of random walks is that they capture the structure of the graph and node properties by randomly visiting neighboring nodes and sampling the graph. In order to mathematically represent these node properties, a GRL algorithm maps node properties in a graph into a low-dimensional space. 

\noindent \begin{definition}
Given an input graph $G = (V, E)$, a graph representation algorithm $f:G \xrightarrow{} \mathbb{R}^d$ maps nodes of the graph into a $d$-dimensional space that captures the properties and closeness of the nodes.
\end{definition}

GRL algorithms are widely used in practice and one of the most employed methods is collecting multiple random walks starting from each node in the graph and converting them to a low dimensional embedding using word2vec \cite{Perozzi2014DeepWalkOL,Grover2016node2vecSF}, a natural language processing (NLP) technique. These node embeddings can be used in several machine learning based graph learning tasks, such as link prediction and node classification since they significantly reduce the complexity of input graph data.

\begin{figure}
\centering
\includegraphics[width=\columnwidth]{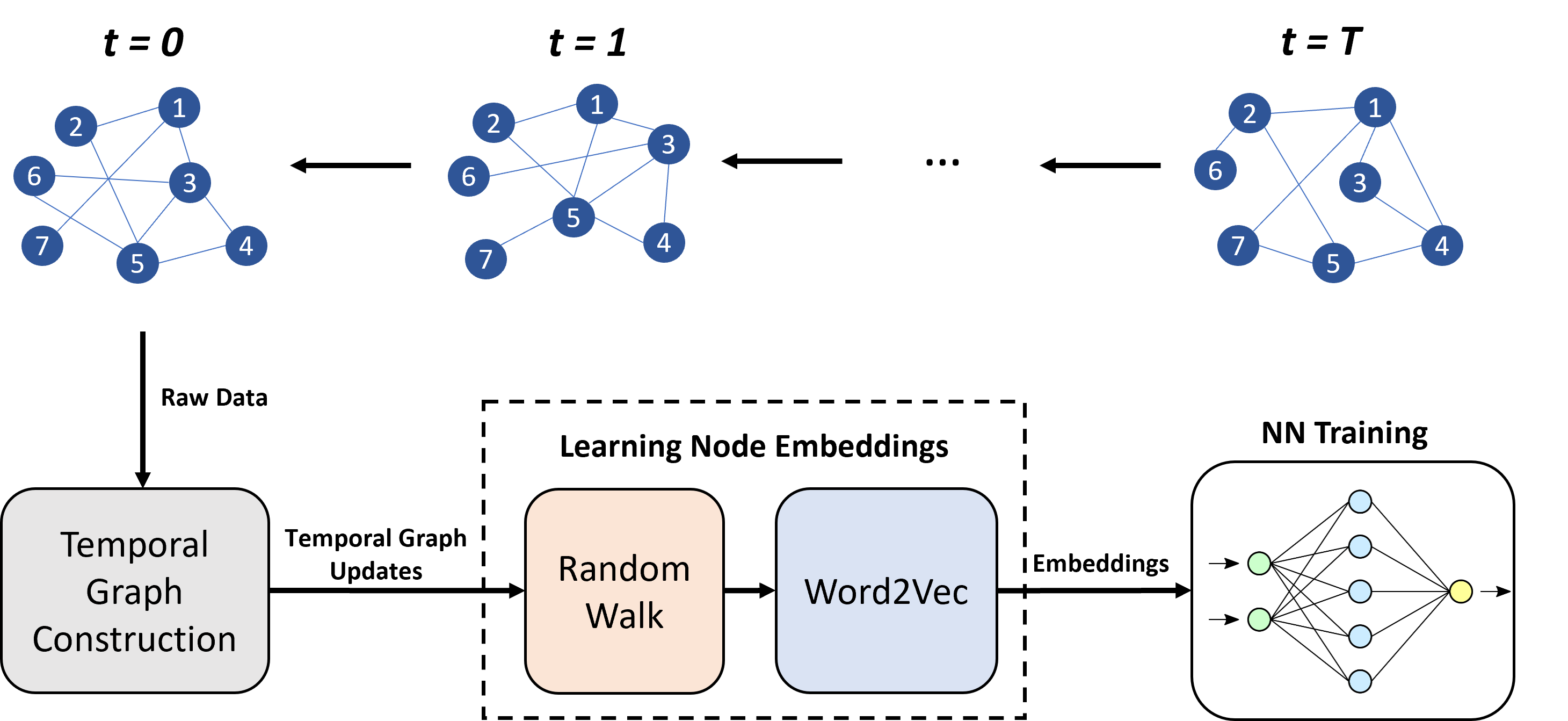} 
\caption{The overview of our proposed temporal graph learning pipeline. The pipeline takes snapshots of a temporally changing graph at each timestamp $t$ and processes the temporal updates using a graph construction step. These updates are then sent to random walk and word2vec steps to map the graph snapshot into a low-dimension embedding space. Finally, the updated graph embeddings are fed to an FNN training step for link prediction or node classification.} 
\label{fig:pipeline}
\end{figure} 

In the proposed temporal implementation, since the input graph is dynamic and temporally updating, we take snapshots of the graph in timestamps and apply the GRL algorithm using the temporal updates coming from each timestamp. As shown in Figure \ref{fig:pipeline}, these updates are sequentially processed and node embeddings are learned, not in one single iteration, but as the graph evolves. In every timestamp, the learned node embeddings are fed to the training model. Therefore, the node embeddings used for training only contain the past and current temporal updates, but not the future ones. For example, in the link prediction task, the embeddings of the nodes that are linked to each other are given to the model as training inputs. However, since the graph is temporally evolving, these node embeddings change over time. Because of this reason, when an edge at timestamp $t$ is given to the training model as an input, the node embeddings only take the graph updates at timestamp $t$ and previous timestamps $\{t-1, t-2, ..., 0\}$ into consideration. In other words, since the graph updates from $t+1$ is not available yet, the node embeddings should not contain this information when the model is being trained at timestamp $t$. This provides a more realistic training approach. Below, we explain how each individual step in the temporal GRL pipeline operates.  

\subsection{Graph Construction}

\begin{figure}
\centering
\includegraphics[width=0.9\columnwidth]{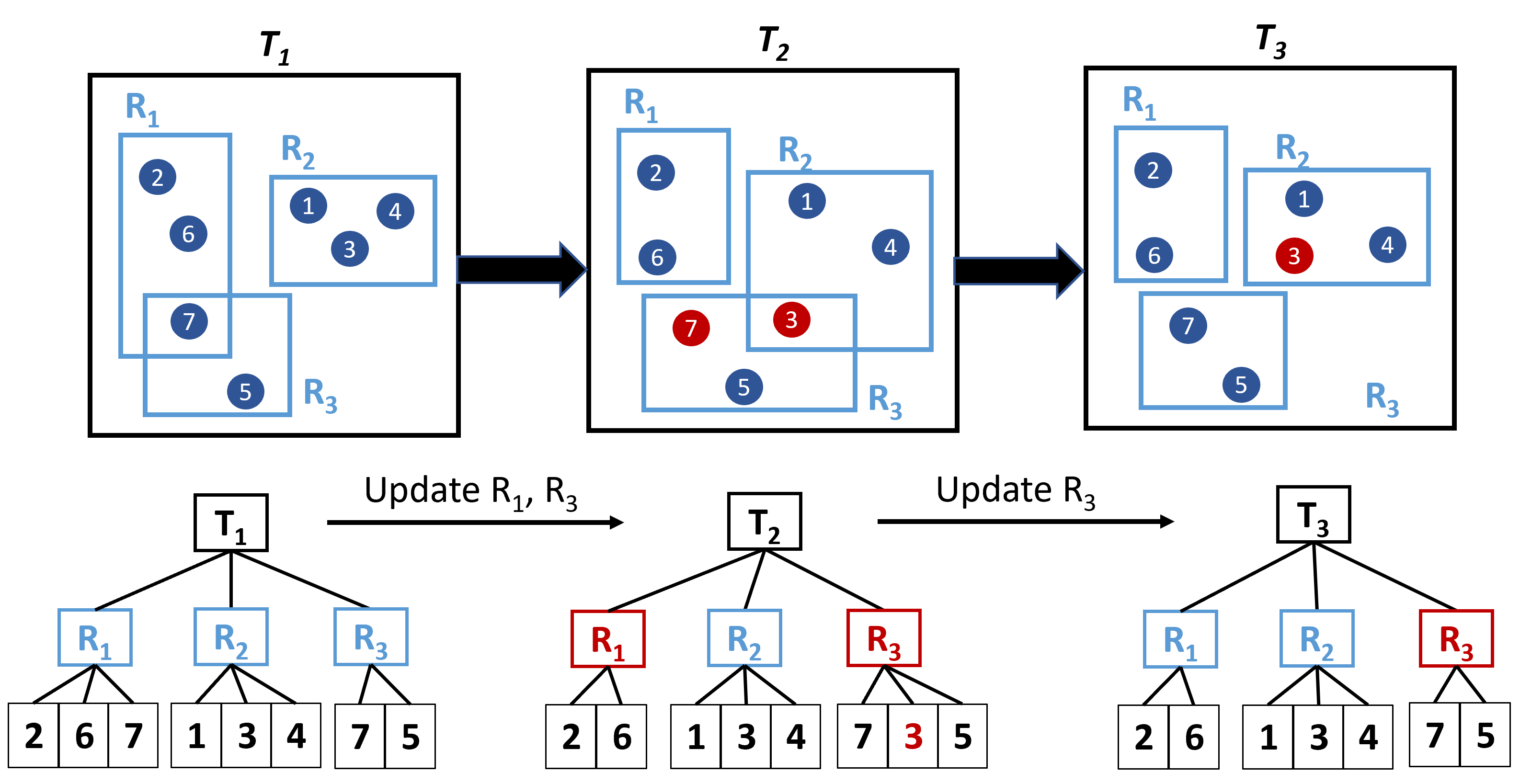}
\caption{An overview of graph construction with R-Trees, which clusters all nodes that are linked to each other in MBRs and only updates a portion of the tree structure as the graph evolves over time to avoid redundant updates.}
\label{fig:construction}
\end{figure}

The pipeline starts with taking raw graph data that contains node ID, edges and temporal information. It first constructs a dynamically updated graph and updates the graph structure with upcoming nodes and edges. The traditional approach is to construct static networks from the temporal snapshots of the graph that cannot handle temporally
changing data. There have been works that attempted to construct temporal graphs more efficiently\cite{Mahmood2018SpatiotemporalAM}. We follow an efficient R-Tree based construction method \cite{Gurevin2021AnEA} that generates and maintains a temporal graph. The proposed method is based on a customized R-Tree based constructor to keep track of all nodes and their interactions. It avoids redundant updates as nodes evolve over time, resulting in a significant reduction in temporal graph construction time compared to the naive construction. 
Although a graph construction step has not been implemented and evaluated in the baseline method, in a temporal setting, only the new updates from the graph should be considered while learning the node embeddings. For this reason, this R-Tree based graph construction method is a suitable technique for keeping track of node interactions between timestamps.

In this step, when the temporal graph data at timestamp $t$ is streamed, all the nodes that form edges with each other are clustered together in minimum bounding rectangles (MBRs), and represented in a tree structure as seen in Figure \ref{fig:construction}. While some nodes change their interactions over time, not every node changes its interactions at each timestamp. Therefore, instead of processing every node, we only update MBRs that contain a node that has changed its interactions. No updates are needed in the tree structure as long as the nodes remain in their MBRs.

\subsection{Random Walks and word2vec}

For random walks, we borrow the idea of EvoNRL \cite{Heidari2020EvolvingNR} that continuously learns embeddings from the
temporally updating graph into low-dimension graph representations using random walks without redundant computations. 
In our setting, although the graph is temporally evolving, not all nodes and edges change in every timestamp. Therefore, it is redundant to reconstruct the random walks for those nodes if they and their neighbors do not change. For this reason, the algorithm maintains a list of random walks for each node. In the next timestamps, in the random walk list, \textit{only the random walks that contain the nodes that are affected by the temporal changes} are reconstructed. This way we (1) maintain a set of random walks that are consistently valid with respect to the graph changes, and (2) eliminate the effect of the random processes by preserving, as much as possible, the original random walks that haven’t been affected by the graph changes.

\begin{figure}
\centering
\includegraphics[width=0.7\columnwidth]{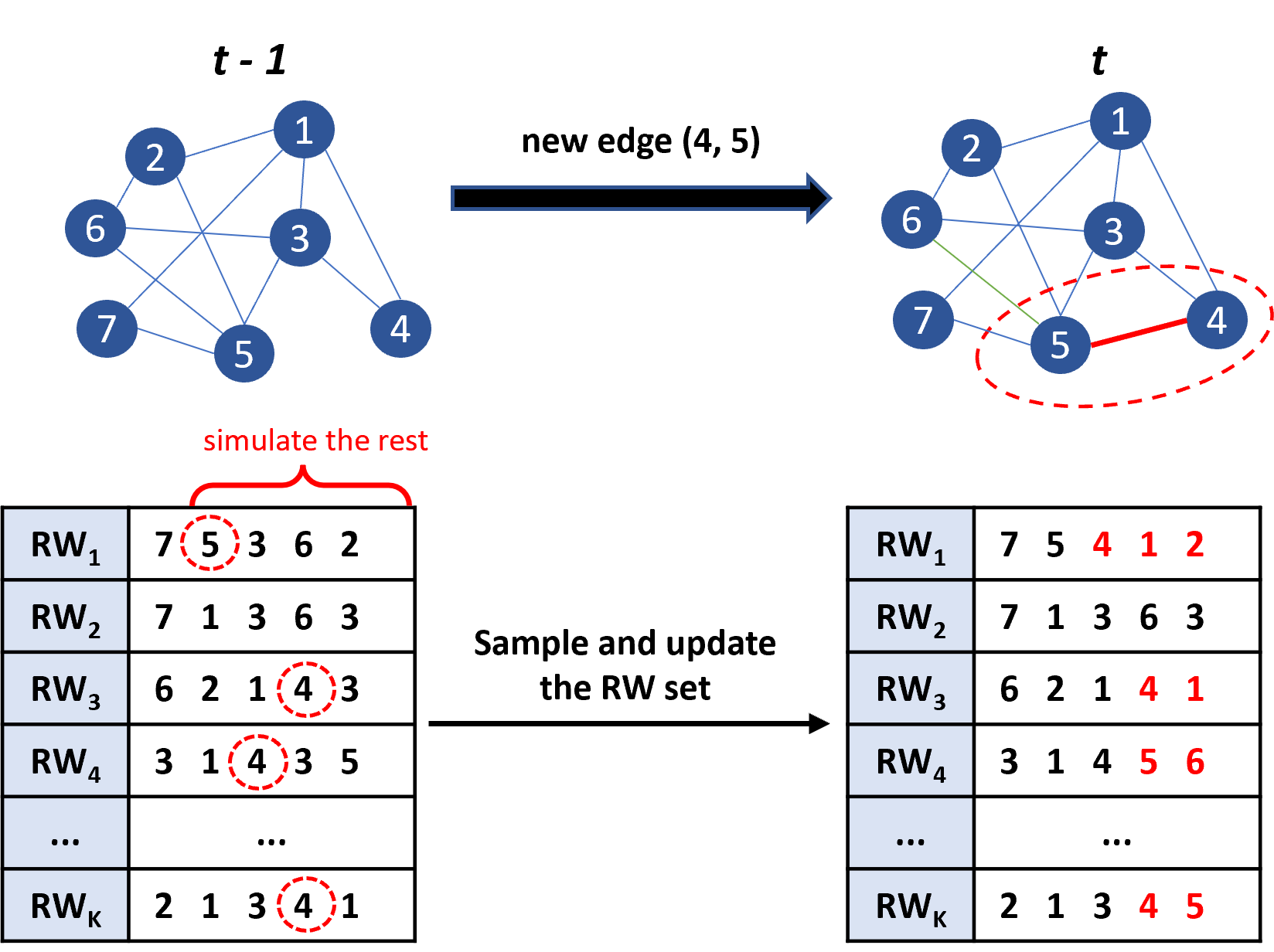}
\caption{Demonstration of a random walk update for an added edge from timestamp $t-1$ to $t$. Only a portion of the random walks including one of the edge nodes are updated.}
\label{fig:rw}
\end{figure}

When the graph temporally changes from one timestamp to another, new edges are formed and some existing edges are removed. When an edge $(u, v)$ is formed (or removed) at timestamp $t$, the immediate neighbors of node $u$ and node $v$ and their interactions with other nodes are consequently affected by this change if new random walks are collected. Instead of collecting new random walks, the random walk list from the previous timestamp $t-1$ is utilized and the random walks containing either node $u$ or $v$ are retrieved and updated. This update only applies to the portion of the random walk after the updated node, i.e. only the rest of the walk that comes after node $u$ or $v$ need to be re-simulated. At each timestamp $t$, after the necessary updates are assigned to the random walk set, the updated random walks are fed to the word2vec model to map the graph structure into a $d$-dimensional space. Figure \ref{fig:rw} shows an example of how this process works. The approach differs from the previous work \cite{Talati2021ADD} that uses word2vec once to capture node embeddings. However, the proposed approach invokes word2vec at each timestamp and collects \emph{temporal} embeddings on the updated graph data.

\subsection{Online FNN Training}

A feed-forward neural network (FNN) is utilized for training on the graph data. Inputs to this step include the node embeddings from the GRL step. Link prediction uses a 2-layer FNN that takes an edge list (concatenation 2 node embeddings) and their corresponding labels that show whether they are present or absent in the graph. Node classification uses a 3-layer FNN that takes node embeddings and their corresponding category labels as inputs.

\begin{figure}[t]
\centering
\includegraphics[width=\columnwidth]{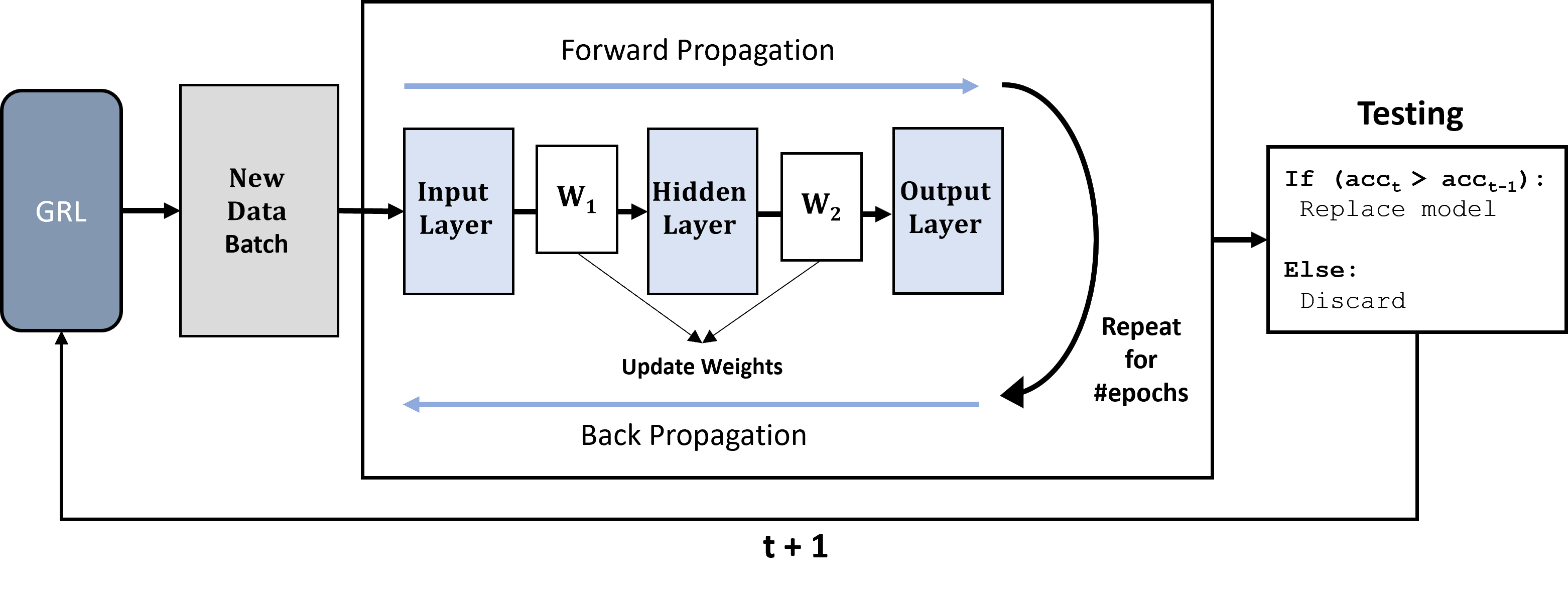} 
\caption{Online training approach demonstrated for a 2-layer FNN. For streaming data batches containing updated node embeddings from the graph representation learning (GRL) algorithm, forward and back propagations are applied on a single instance in each online iteration.}
\label{fig:online}
\end{figure}

In our setting, training data is sequentially streamed in as the graph evolves in each timestamp. Therefore, we adopt an \emph{online learning} strategy to train the FNN at run-time \cite{Sahoo2018OnlineDL, Yoon2018LifelongLW, Lee2016DualMemoryDL}. Figure \ref{fig:online} depicts our approach. When a new data batch arrives from the GRL step, it is fed into the training model where forward and back propagation steps are applied on the same batch for consecutive iterations.
Here, our intuition is that temporal batches that are large in size already contain sufficient information about the graph since they include node embeddings that capture graph history from from all previous timestamps $\{0, 1, ..., t\}$. Hence, by collecting enough updates our training approach tolerates slight over-fitting for investigating the same batch or additional iterations before moving on to the next one. Moreover, this approach can have some performance advantages such as reducing the memory overheads related to data loading on the internal memory for training. In future work, more advanced online learning techniques can be adopted, such as evolving the FNN model depth from simple to complex over time\cite{Sahoo2018OnlineDL}.

\section{Exploiting Parallelization in the FNN Pipeline}\label{sec:parallel_mm}

\begin{figure}
\includegraphics[width=\columnwidth]{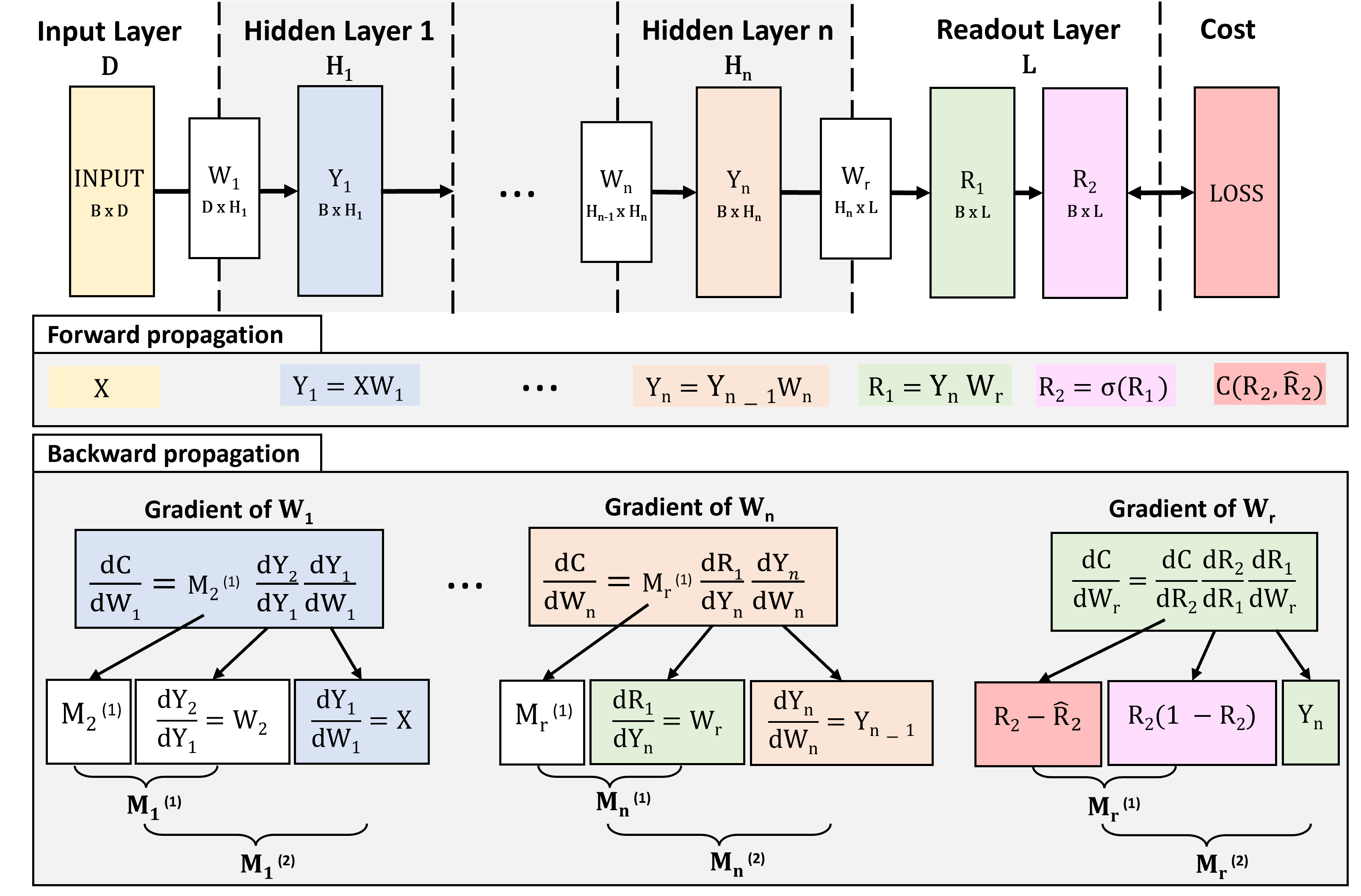} 
\caption{Breakdown of the individual matrix operations in the forward and backward propagation during the training of an FNN with $n$ hidden layers. $D$ is the node embedding size and $B$ is the batch size. \{$H_1, H_2, ..., H_n$\} are the sizes of the hidden layers \{$Y_1, Y_2, ..., Y_n$\}. $L$ represents the number of labels of the nodes.}
\label{fig:fnn_pipeline}
\end{figure}

Talati \etal \cite{Talati2021ADD} proposed parallel implementations to accelerate the performance of the random walk and word2vec steps. However, it identified that FNN training significantly dominates the execution time of the end-to-end workload and left acceleration of training as future work. In this paper, we provide a more in-depth analysis of the FNN training to identify the performance bottlenecks and propose parallelization techniques for their acceleration.

\subsection{In-Depth Analysis of the FNN Pipeline}

We first break the FNN training into its components in the finest granularity, which are the individual matrix operations in each step. Figure \ref{fig:fnn_pipeline} demonstrates the training pipeline and each step taken in the forward and backward propagation for an FNN with $n$ hidden layers. In forward propagation, the input batch $X$ of size $D$ (node embedding dimension size) and $B$ (batch size) is fed to the input layer and propagates through $n$ hidden layers $\{Y_1, Y_2, ..., Y_n\}$ with sizes $\{H_1, H_2, ..., H_n\}$, respectively. Each hidden layer $Y_i$ has its own corresponding weight matrix $W_i$ and can be computed as $Y_i = Y_{i-1} \times W_i$. The final layer is the readout layer $R_1$ of size $L$ (the number of labels), with its weight matrix $W_r$. It is followed by an activation function that produces $R_2$. Finally, the cost function is applied to compute the loss between the final layer $R_2$ and the correct labels $\hat{R_2}$. This completes the forward propagation stage. After the loss is computed, backward propagation starts to calculate the gradient for each weight matrix $W_i$ for $i \in \{0, 1, ..., n, r\}$. This is done by taking the derivative of the cost function $C$ with respect to the weight matrix using the chain rule. After computing the gradient of $W_r$, the gradient computation of the subsequent layers only consist of two main matrix multiplications: $M_{i}^{(1)}$ and $M_{i}^{(2)}$ since some portion of the chain derivatives are already computed from the previous layer. After the backward propagation step is completed, weight matrices are updated with their own corresponding gradients.

\subsection{Parallelization of Low-Dimensional Kernels}

\begin{figure}[t]
  \centering
  
  \subcaptionbox{Inner Product \label{fig:inner}}{\includegraphics[width=0.35\columnwidth]{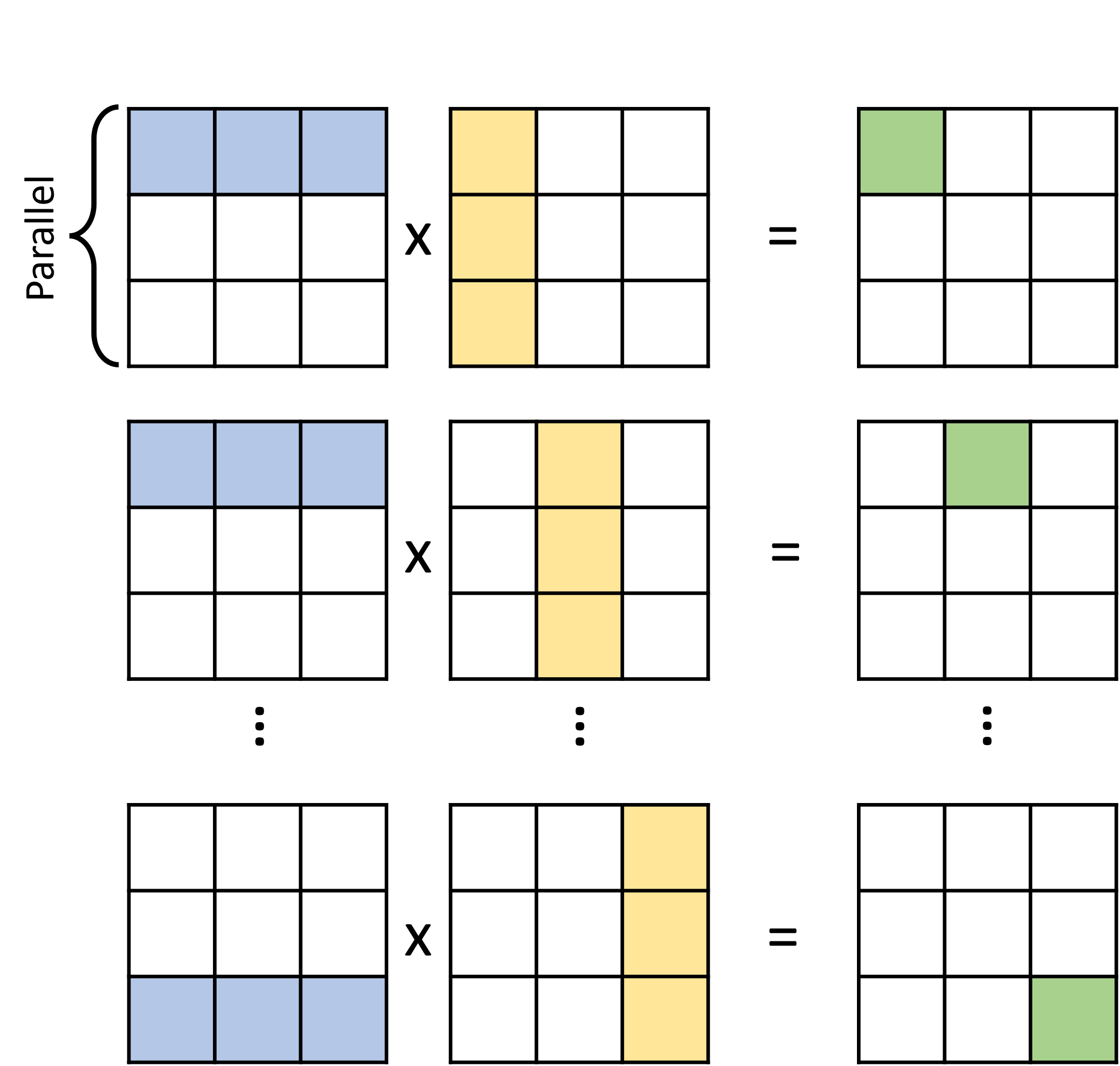}}\hspace{5mm}
  \subcaptionbox{Outer Product \label{fig:outer}}{\includegraphics[width=0.35\columnwidth]{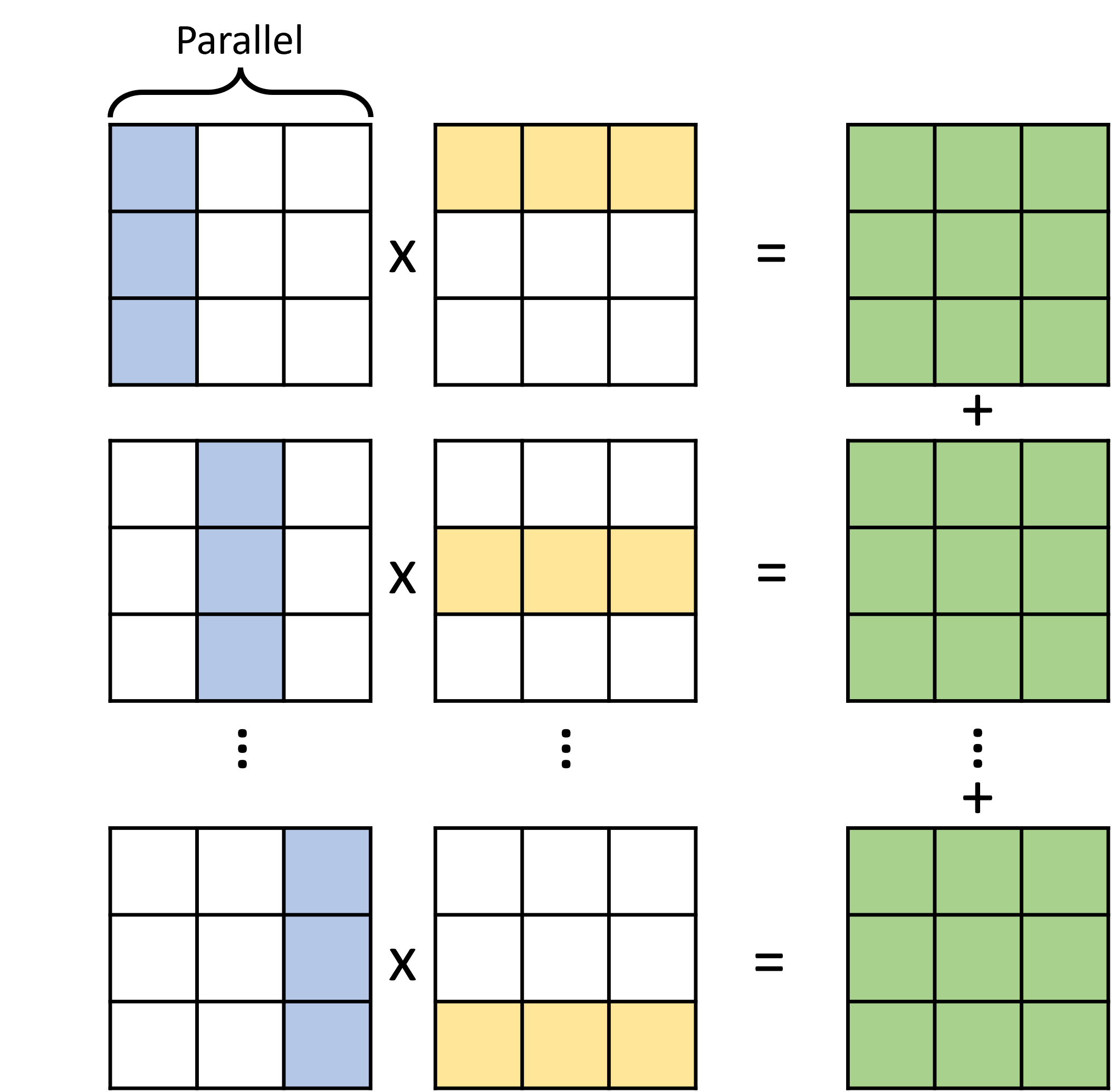}}\hspace{5mm}

  \subcaptionbox{Row-Wise Product \label{fig:row}}{\includegraphics[width=0.35\columnwidth]{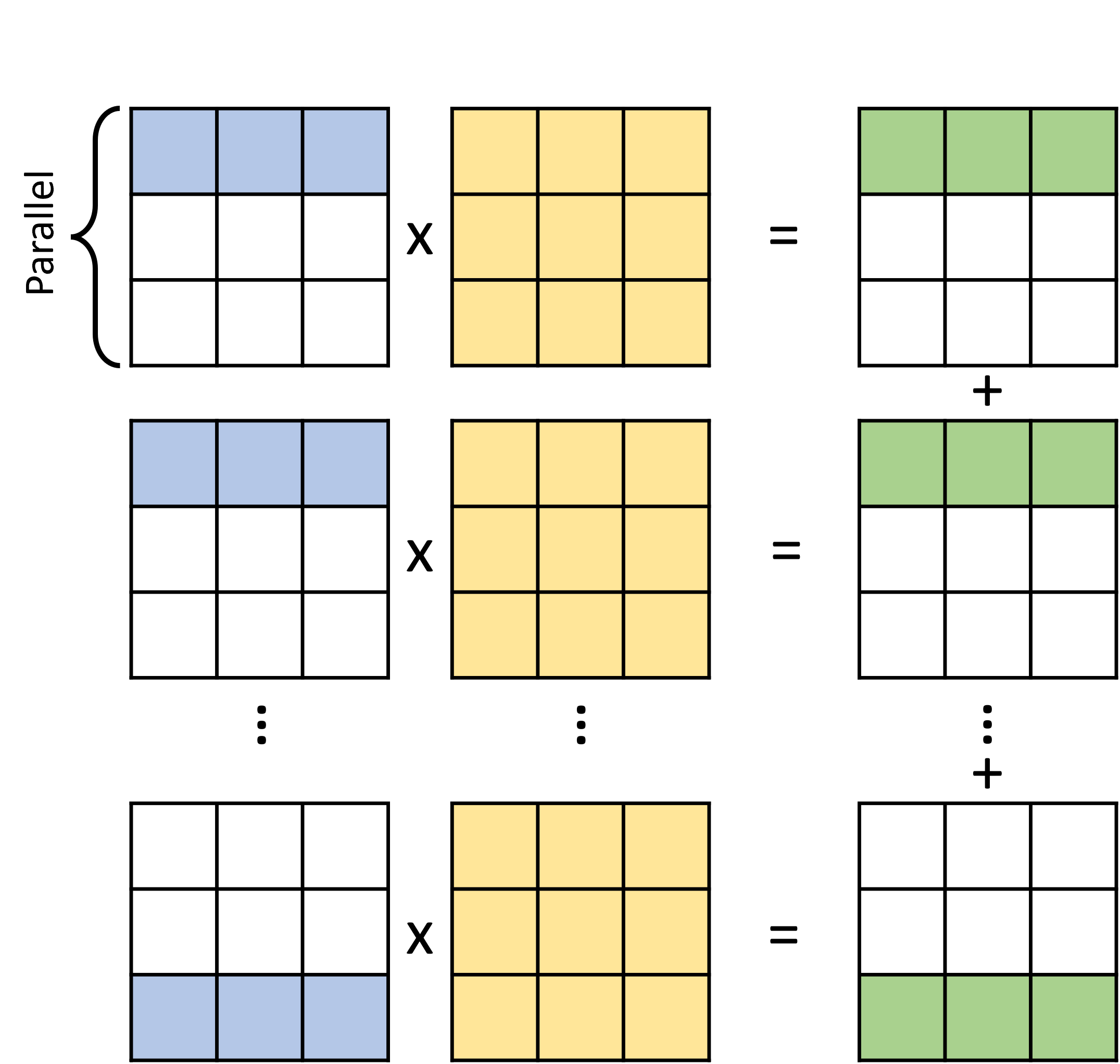}}\hspace{5mm}
  \subcaptionbox{Column-Wise Product \label{fig:column}}{\includegraphics[width=0.35\columnwidth]{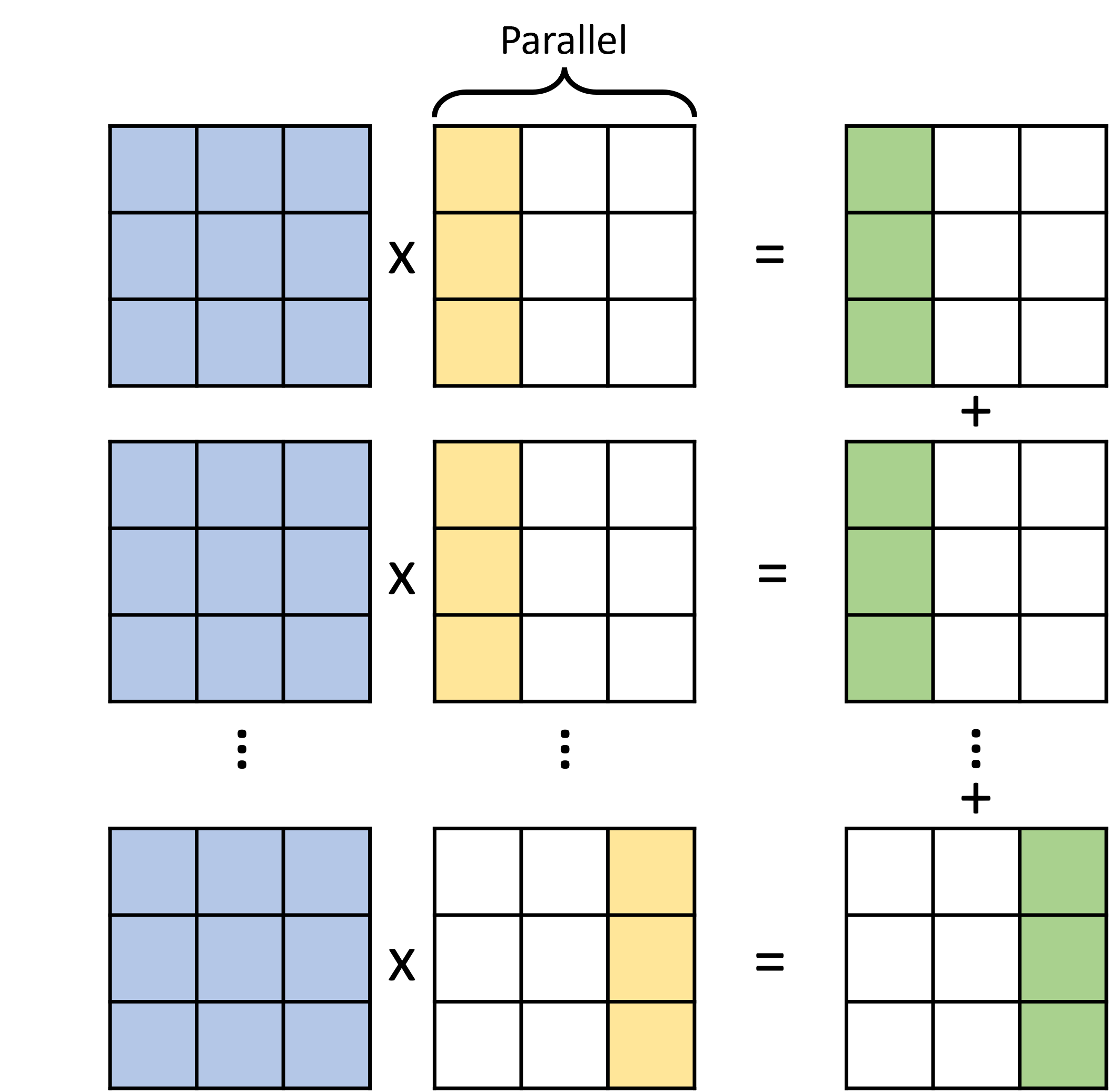}}\hspace{5mm}

  \caption{Four different approaches for computing dense-dense matrix multiplications (DDMMs) in parallel. The elements of the first and second input matrices are shown in blue and yellow color, respectively. The elements of the output matrix are shown in green color.  }
\label{fig:MM}
\vspace{-2mm}
\end{figure}

The matrix operations in the FNN pipeline are the main bottlenecks that contribute to the overall training cost. It has been identified that there is limited parallelism potential for matrix-multiplications (MMs) due to the small and different matrix sizes in the studied pipeline \cite{Talati2021ADD}. In this section, we discuss how we can exploit parallelism for these low-dimensional kernels by using several state-of-the-art approaches.

One way of accelerating MMs is to distribute the operations in rows or columns of the input matrices among multiple cores and process them in parallel. In a matrix multiplication, $X \times Y = Z$, each of the input matrices $X$ and $Y$ can be accessed in either a row-wise or column-wise manner. Therefore, the matrix operations can be performed in four different ways: (1) inner product, (2) outer product, (3) row-wise product and (4) column-wise product as shown in Figure \ref{fig:MM}.

\noindent \textbf{Inner Product.} This is the most commonly used MM approach. As shown in Figure \ref{fig:inner}, it reads a row of the first input matrix $X$ and a column of the second input matrix $Y$, and performs an index-matching dot product between two vectors. As a result, one cell of $Z$ is updated. This MM technique can be performed by distributing the rows of $X$ among different cores. One disadvantage is that each row of $X$ and each column of $Y$ needs to be read multiple times to update a single row of $Z$. Even though there is high locality for the repeated reads for the rows of $X$, they can still be costly. For this reason, this approach works better when the input matrices are small. On the other hand, each cell of $Z$ is read and updated once. This allows for different cores to write in different cells of $Z$. This makes it suitable for parallelism and provides better scaling when the output matrix is large.

\noindent \textbf{Outer Product.} As shown in Figure \ref{fig:outer}, outer product \cite{Bulu2008OnTR, Geijn1997SUMMASU} is performed between a column of the first matrix $X$ and a row of the second matrix $Y$. The result produces a partial sum for all cells of $Z$. These partial sums are accumulated in each cell after each multiplication. For parallelization, columns of the first matrix $X$ are distributed among different cores. The advantage of outer product is that the rows and columns of the input matrices are only read once. This makes outer product a good option when the input matrices are large. On the other hand, the output matrix $Z$ needs to be updated atomically that impact performance scaling at higher core counts. However, outer product can still yield acceleration if both input matrices are very large and the output matrix is small.

\noindent \textbf{Row-Wise Product.} As shown in Figure \ref{fig:row}, row-wise product \cite{Gustavson1978TwoFA} takes a single row of the first input matrix $X$ and multiplies it with the rows of the second matrix $Y$. Each multiplication outputs a partial sum for a row of $Z$ and these partial sums are accumulated after each multiplication. Compared to inner product, row-wise product reads the single row of $X$ once and performs the multiplication by reading all of the rows of $Y$. For parallelization, it is similar to inner product and the rows of the first matrix are distributed among different cores, and output matrix updates do not require atomic operations. Row-wise product yields better performance if the second input matrix is small, and thus preferred to inner product since there are less number of reads to the first matrix.

\noindent \textbf{Column-Wise Product.} As shown in Figure \ref{fig:column}, this approach takes a single column of the second input matrix $Y$ and multiplies it with all the columns of the first input matrix $X$. This produces partial sums for a single column of the output matrix. This approach is similar to row-wise, and has the same number of data reads and memory requirement. However, the columns of the second matrix are distributed among cores for paralellization. For this reason, depending on the row and column counts of the input matrices, column-wise or row-wise product might have better performance than each other.

In Section \ref{sec:exp}, each of these parallelization techniques are applied to the matrix multiplications of the evaluated FNNs. Note that the embeddings that are input to the FNN are \textit{dense}. Therefore, in this paper we only explore and evaluate dense-dense MMs. However, in future work, these input embeddings and the weight matrices can be sparsified, which has previously been explored through feature sparsification in GNNs\cite{Chen2018FastGCNFL, Chen2021AUL,You2022EarlyBirdGG}, and weight pruning techniques in DNNs \cite{Zhang2018ASD,gurevin2020enabling}. This approach can unlock sparse-sparse matrix multiplications (SpGEMM) and further accelerate the FNN pipeline.
\section{Methodology}

We characterize our implementations on two CPU platforms. For the temporal graph processing, we conducted our experiments on an Intel Core i7-7820HK (4 physical cores, 8 logical cores at 2.90 GHz) machine with a 32GB memory an 8GB last-level cache (LLC). For the parallelization of the FNN pipeline, we use a 20-core Intel Xeon E5-2650 v3 multicore CPU with 2 sockets and 10-cores per socket. The machine has 512GB memory and a 25MB LLC.

\noindent \textbf{Datasets.} We use four real-world temporal graph datasets for our evaluation: wiki-talk \cite{Leskovec2014SNAPD,Paranjape2017MotifsIT,Cunningham2019CreatorGI}, ia-email \cite{Rossi2015TheND, Shetty2004TheEE} and stackoverflow \cite{Leskovec2014SNAPD,Paranjape2017MotifsIT} datasets for link prediction, and brain dataset \cite{Xu2019SpatioTemporalAR, Preti2017TheDF} for node classification. We divide the graphs in snapshots to process them temporally. The details of the datasets can be found in Table \ref{tab:dataset}.

\begin{table}[]
\caption{Parameters of the temporal graphs used for experiments.}
\label{tab:costtable}
\begin{adjustbox}{width=\columnwidth}

\begin{tabular}{c|c|c|c|c}
\hline
\textbf{Task    }            & \textbf{Dataset}       & \textbf{\#Nodes}   & \textbf{\#Edges}    & \textbf{\#Timestamps} \\ \hline\hline
Link Prediction     & ia-email      & 87,274    & 1,148,072  & 2,244        \\ \hline
Link Prediction     & wiki-talk     & 1,140,149 & 7,833,140  & 16,000       \\ \hline
Link Prediction     & stackoverflow & 6,024,271 & 63,497,050 & 125,000      \\ \hline
Node Classification & brain         & 5,000     & 1,955,488  & 10           \\ \hline
\end{tabular}
\label{tab:dataset}
\end{adjustbox}
\end{table}

\noindent \textbf{Python Implementation of Run-time GRL.} For temporal graph processing, we implement a Python framework that contains: (1) graph construction, (2) temporal random walks \& word2vec, (3) data preparation, and (4) online FNN training and testing steps. We use \texttt{Python 3.6-64} in our implementation. For graph construction, we implemented an R-Tree based construction based on \cite{Gurevin2021AnEA}. For random walks, we use an open source implementation\footnote{\url{https://github.com/farzana0/EvoNRL}} by \cite{Heidari2020EvolvingNR} and use the word2vec model from Python's \texttt{gemsim3.8.1} library. For the data preparation and training codes, we implement a Python version of the \texttt{C++} implementation\footnote{\url{https://github.com/talnish/iiswc21_rwalk}} of \cite{Talati2021ADD}. The downstream FNN is implemented using the PyTorch library. For node classification, we use batch size $B=512$, embedding size $D=64$, hidden layer sizes $H_1=128$ and $H_2 = 256$ and output size $L=10$. For link prediction we use $B=1024$, $D=8$, $H_1=128$ and $L=1$.

\begin{table}[t]
\caption{Sizes of the main matrix multiplications in link prediction and node classification FNNs.}
\label{tab:matrix_sizes}
\begin{adjustbox}{width=\columnwidth}
\begin{tabular}{|c|c|cl|cl|}
\hline
\textbf{Task}                          & \textbf{\begin{tabular}[c]{@{}c@{}}MM\end{tabular}} & \multicolumn{2}{c|}{\textbf{Matrix   1}} & \multicolumn{2}{c|}{\textbf{Matrix   2}} \\ \hline
\multirow{5}{*}{Link   Prediction}     & $Y_1$                                                                       & $X^T$               & ($1024$, $16$)            & $W_1$               & ($16$, $128$)             \\
                                       & $R_1$                                                                       & $Y_1$               & ($1024$, $128$)           & $W_r$               & ($128$, $1$)              \\
                                       & $M_{r}^{(2)}$                                                                   & $Y_{1}^{T}$              & ($128$, $1024$)           & $M_{r}^{(1)}$            & ($1024$, $1$)             \\
                                       & $M_{1}^{(1)}$                                                                    & $M_{r}^{(1)}$            & ($1024$, $1$)             & $W_r$               & ($1$, $128$)              \\
                                       & $M_{1}^{(2)}$                                                                    & $X$                & ($16$, $1024$)            & $M_{1}^{(1)}$           & ($1024$, $128$)     \\ \hline
\multirow{8}{*}{Node   Classification} & $Y_1$                                                                       & $X$                & ($512$, $64$)             & $W_1$               & ($64$, $256$)             \\
                                       & $Y_2$                                                                       & $Y_1$               & ($512$, $256$)            & $W_2$               & ($256$, $128$)            \\
                                       & $R_1$                                                                       & $Y_2$               & ($512$, $128$)            & $W_r$               & ($128$, $1$)             \\
                                       & $M_{r}^{(2)}$                                                                    & $Y_{2}^{T}$              & ($128$, $512$)            & $M_{r}^{(1)}$            & ($512$, $10$)             \\
                                       & $M_{2}^{(1)}$                                                                    & $M_{r}^{(1)}$            & ($512$, $10$)             & $W_{r}^{T}$              & ($10$, $128$)             \\
                                       & $M_{2}^{(2)}$                                                                    & $Y_{1}^{T}$              & ($256$, $512$)            & $M_{2}^{(1)}$            & ($512$, $128$)            \\
                                       & $M_{1}^{(1)}$                                                                   & $M_{2}^{(1)}$            & ($512$, $128$)            & $W_{2}^{T}$              & ($128$, $256$)            \\
                                       & $M_{1}^{(2)}$                                                                    & $X_T$               & ($64$, $512$)             & $M_{1}^{(1)}$           & ($512$, $256$)            \\ \hline
\end{tabular}
\end{adjustbox}
\end{table}

\noindent \textbf{C++ Implementation of the FNN Pipeline.}
To exploit parallelism for the matrix multiplication operations in the FNN pipeline, we have implemented a \texttt{C++} framework that performs an entire training iteration with custom implementations of the forward and backward propagation steps. The framework implements multi-threaded inner, outer, row-wise and column-wise parallelization techniques for each matrix kernel. The \texttt{pthreads} library is used to utilize the available core counts on the machine, and are compiled using the \texttt{g++ v6.4.1} compiler. Table \ref{tab:matrix_sizes} lists all the evaluated MMs with their sizes for link prediction and node classification FNNs. We exclude $M_{r}^{(1)}$ since it is always an element-wise multiplication. The rest of the operations shown in Table \ref{tab:matrix_sizes} are regular matrix multiplications.

\begin{figure}
\includegraphics[width=\columnwidth]{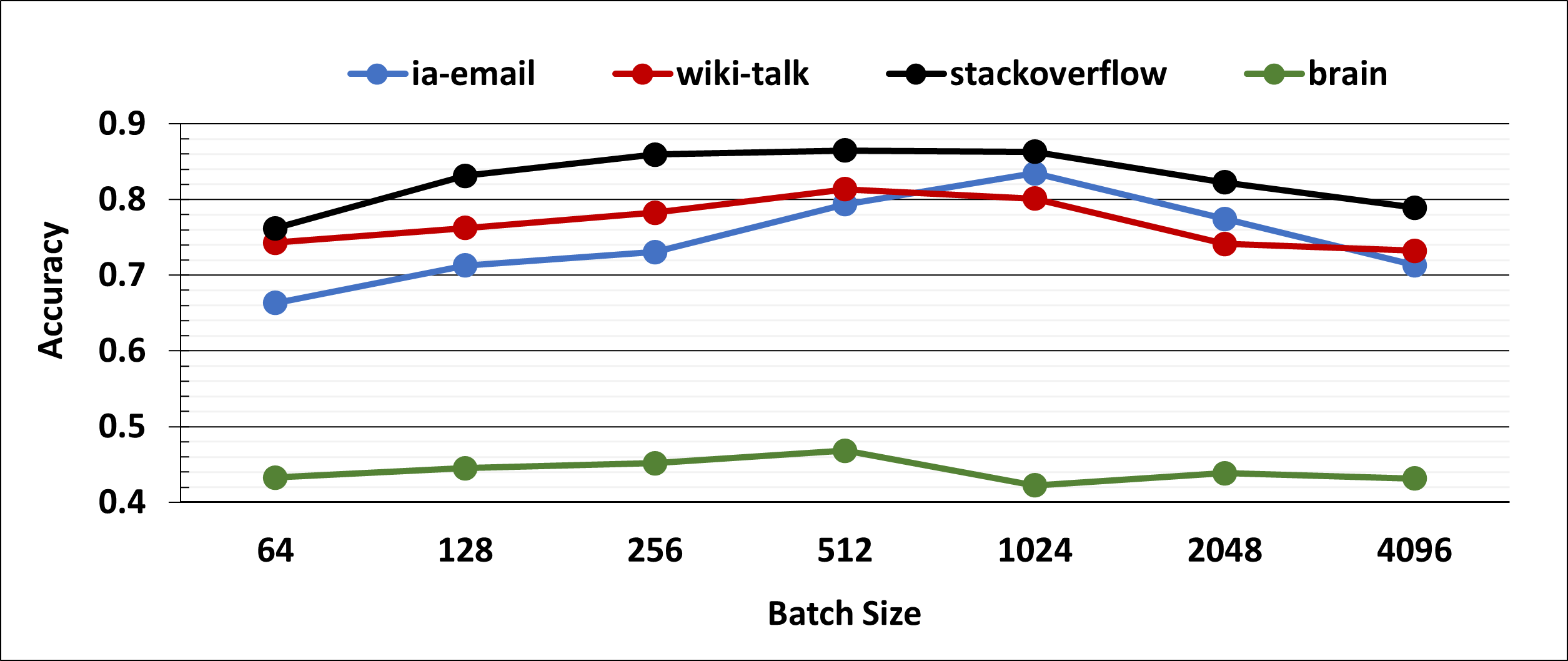} 
\centering
\caption{{Accuracy of temporal training with different batch sizes.}}
\label{fig:acc_batch}
\end{figure}

\section{Experimental Evaluation}\label{sec:exp}
\subsection{Temporal Processing Pipeline}

\begin{figure}[b]
\includegraphics[width=\columnwidth]{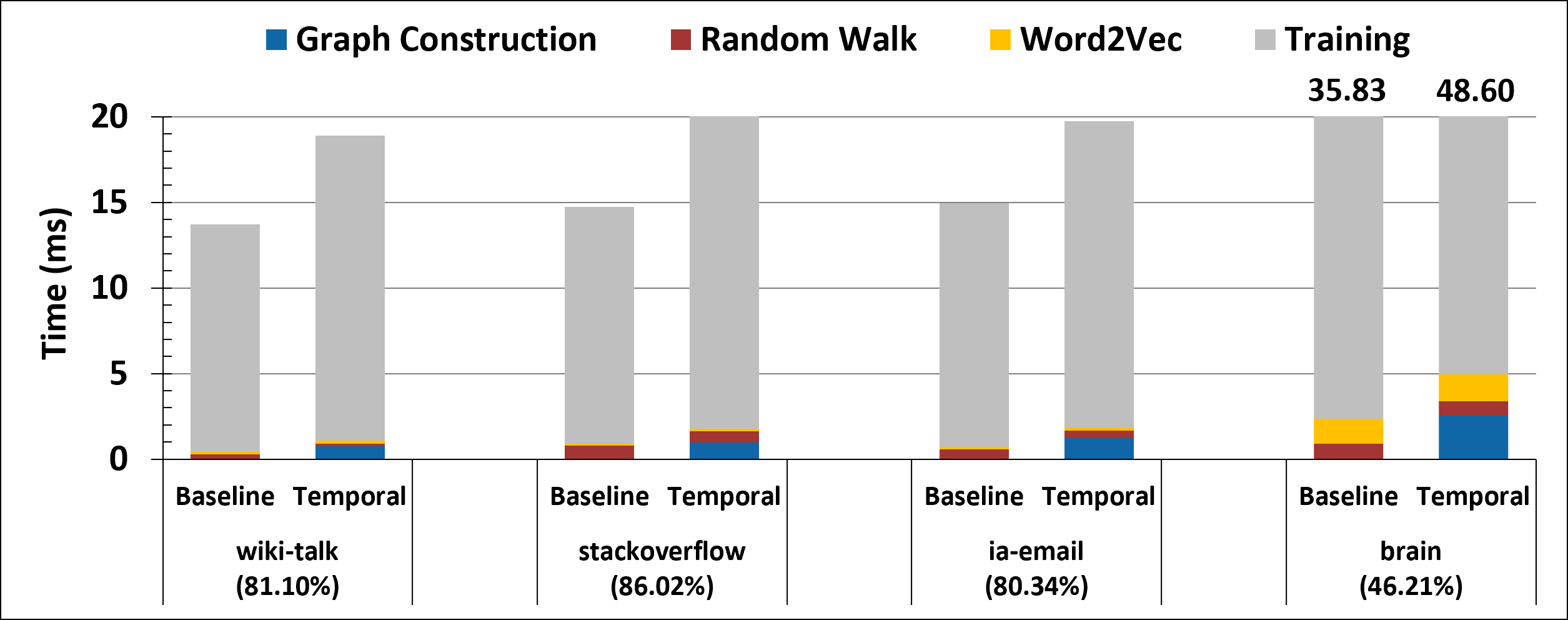}
\centering
\caption{Execution time overheads of the proposed temporal graph learning pipeline and comparison with the prior work \cite{Talati2021ADD} for 4 temporal graph datasets. The final accuracy at the end of training is reported for each dataset.}
\label{fig:temporal_pipeline}
\end{figure}

\begin{figure*}[th]
\begin{subfigure}{\textwidth}
  \centering
  \includegraphics[width=\linewidth]{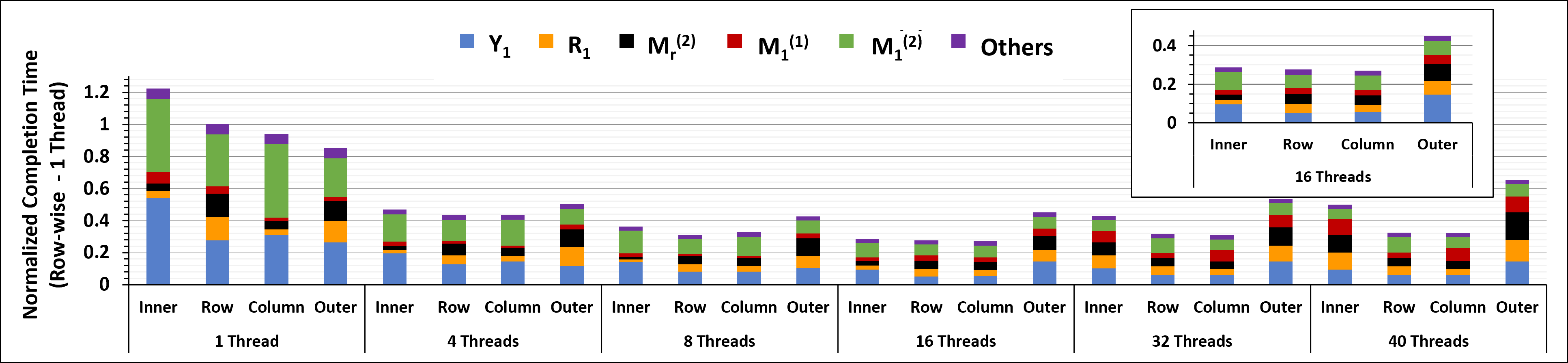}  
  \caption{Link Prediction.}
  \label{fig:link_matrix}
  \vspace{5mm}
\end{subfigure}

\begin{subfigure}{\textwidth}
  \centering
  \includegraphics[width=\linewidth]{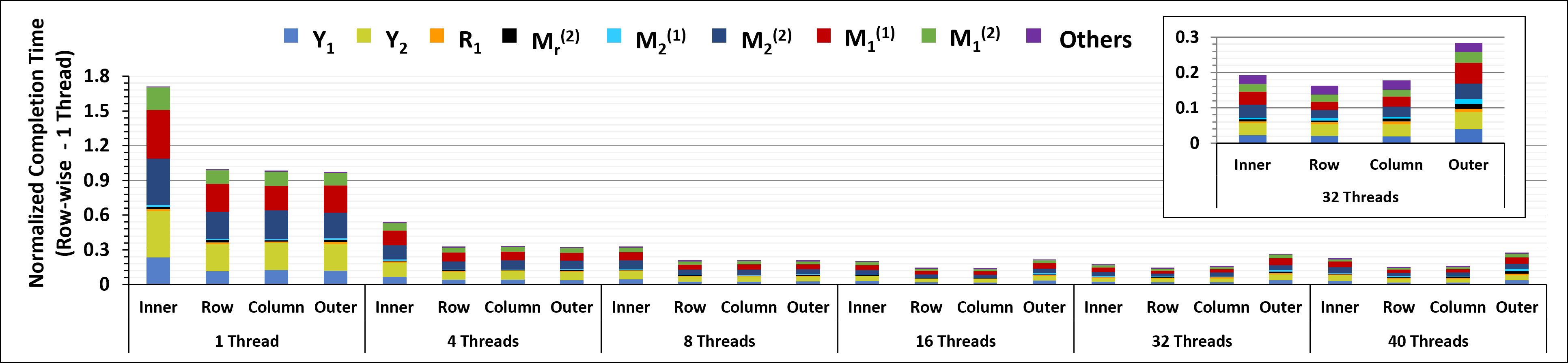}  
  \caption{Node Classification.}
  \label{fig:node_matrix}
\end{subfigure}
\caption{The performance evaluation of the matrix multiplications in the training pipeline of link prediction (wiki-talk) and node classification (brain) FNNs which are parallelized using inner, outer, row-wise and column-wise product algorithms. The scaling for each  is shown for 1, 4, 8, 16, 32 and 40 threads. }
\label{fig:matrix_mul_fig}
\end{figure*}

{Figure \ref{fig:acc_batch} shows the accuracy of the temporal training with different batch sizes for each dataset: ia-email, wiki-talk and stackoverflow for the link prediction task and brain for the node classification task. The link prediction datasets have the best accuracy with a batch size of 1024, which is 83.45\%, 80.1\% and 86.31\% for ia-email, wiki-talk and stackoverflow, respectively. 
The link prediction accuracy is lower for smaller batches because the online training strategy causes slight over-fitting. On the other hand, with larger batch sizes the model accuracy degrades. This is because larger batches cannot capture sufficient temporal information about the graph as some temporal information of the graph is lost when processing larger time windows. The performance of the node classification task on the brain dataset is not significantly affected by different batch sizes since brain dataset has less number of timestamps (i.e. less temporal data). However, a batch size of 512 on brain dataset yields the highest accuracy. Therefore, we fix the batch size to 1024 for link prediction datasets, and 512 for node classification dataset for the rest of the experiments. }

Figure \ref{fig:temporal_pipeline} depicts our experimental results for the implementation of the temporal graph learning pipeline and its comparison with the baseline \cite{Talati2021ADD}. We measured the execution time of different steps for one temporal batch in the pipeline: temporal graph construction, temporal random walk, word2vec and finally training for node classification or link prediction depending on the dataset. In comparison, we measured the execution time of the baseline pipeline for the same size of data. The accuracy for each dataset is reported in Figure \ref{fig:temporal_pipeline}.

Overall, brain dataset takes the most amount time due to the density of temporal updates in the dataset as well as the node classification task having a 2-FNN layer with more matrix kernels. Since the graph construction is not implemented in the baseline, there is an additional overhead for temporal graph construction in our pipeline, which is the most costly operation after the training step. Overall, in comparison to the baseline, the proposed implementation's execution time is comparable.

In conclusion, the experiments show the breakdown of all the steps in the temporal pipeline and verify that FNN training is the performance bottleneck, which confirms the findings of the baseline method \cite{Talati2021ADD}. In the next section we show the breakdown of the FNN training, and propose performance acceleration using parallelization of the low-dimensional kernels.

\subsection{Exploiting Parallelism in FNN Pipeline}  
Figure \ref{fig:matrix_mul_fig} shows the performance analysis for the breakdown of one iteration in the training pipeline. We show the performance of each matrix multiplication kernel in the forward and back-propagation stages using the row-wise, column-wise, inner and outer matrix multiplications.
The results are shown for increasing thread counts from 1 to 40 threads. The sizes of each individual matrix can be found in Table \ref{tab:matrix_sizes}. 

Figure \ref{fig:link_matrix} shows the results for the 2-layer FNN for the link prediction task. All timing measurements are normalized based on the row-wise multiplication at 1 thread. Here, we mainly show the performance of the main matrix multiplications and group the rest of the operations in the pipeline (e.g. sigmoid function, cost function, vector subtractions, etc.) as ``Others". As it can be seen, $Y_1$ and $M_{1}^{(2)}$ are the most expensive matrix operations, followed by $R_1$ and $M_{r}^{(2)}$ in the link prediction FNN. Overall, row-wise and column-wise both have the best performance scaling at $16$ threads. We observe $\sim$$2.5\times$ speedup compared to the single thread execution of row-wise.

Figure \ref{fig:node_matrix} shows the results for the 3-layer FNN for the link prediction task. All  timing measurements are normalized with the row-wise multiplication at 1 thread. Since the matrix sizes are relatively larger in this FNN, we observe better performance scaling for all kernels. At 32 threads, row-wise product shows the best performance scaling with a $6.6\times$ speedup compared the row-wise using 1 thread.

In the above experiments, the same parallelization technique is used for all matrix kernels in an FNN pipeline. 
However, we observe that different parallelization techniques perform better for each individual matrix kernels.
For example, for the link prediction FNN in Figure \ref{fig:link_matrix}, even though row-wise has the best performance for $Y_1$, it is not the case for $M_{r}^{(2)}$ since inner-wise product has the best performance. 
Further performance acceleration is possible by combining different parallelization techniques at the kernel granularity.
We plan to explore such fine-grain methods as future work.

\begin{figure}%
    \centering
    \subfloat[\centering Link Prediction (column-wise). ]{{\includegraphics[width=0.45\columnwidth]{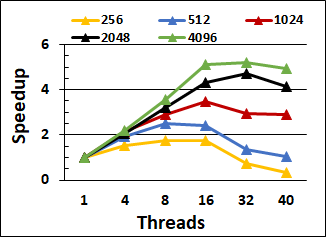}\label{fig:link_batch} }}
   \hspace{-1mm}
    \subfloat[\centering Node Classification (row-wise).]{{\includegraphics[width=0.45\columnwidth]{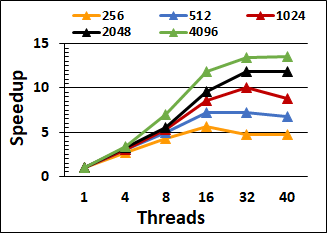}\label{fig:node_batch} }}
    \caption{{Performance scaling of the best MM strategy with different batch sizes at varying thread counts.}}
    \label{fig:batch_fig}
\end{figure}

{Figure \ref{fig:batch_fig} shows the scaling of link prediction (wiki-talk) and node classification (brain) training with different batch sizes with increasing thread counts. For both tasks, the best MM strategy is used from the experiment in Figure \ref{fig:matrix_mul_fig}, that is column-wise for link prediction and row-wise for node classification. Link prediction achieves $5\times$ speedup with a batch size of 4096 at 16 threads. Node classification achieves a higher speedup than link prediction, $13.5\times$, with a batch size of 4096 at 40 threads. This is due to the larger matrix sizes in node classification training pipeline. While larger batch sizes enable better performance scaling for matrix multiplications, larger batch sizes might have negative impact on accuracy depending on the temporal characteristics of the dataset, as previously shown in Figure \ref{fig:acc_batch}.}

\section{Discussion and Future Work}

In this paper, we consider dense input embedding and weight matrices in the FNN during training and inference. We have shown the performance scaling with different dense-dense matrix multiplication parallelization techniques. In graph neural networks, sparsification of the input embedding matrix \cite{Chen2018FastGCNFL, Chen2021AUL,You2022EarlyBirdGG} and weight matrices \cite{Zhang2018ASD,gurevin2020enabling} is a common approach for accelerating the training and inference of models. These sparsification approaches can be applied to our proposed temporal training pipeline, which will result in sparse-dense and/or sparse-sparse matrix multiplications in forward and backward propagation stages. Through parallelization of sparse-dense and sparse-sparse matrix multiplications, further acceleration can be achieved.

Other than exploiting the redundancy in embedding and weight matrices, redundancy between temporal timestamps can also be exploited for further acceleration. The redundancy between consecutive timestamps can be explored by observing the learning trajectory of the FNN during training. One possible way is to monitor the loss of temporal batches after forward propagation. The loss is an interpretation of how well the model performs for a particular set of data. A high loss value on a batch indicates that the model has not yet learned enough from the previous batches to be able to make a good prediction for that particular batch. Therefore, it is an indicator of better learning opportunity for the model. On the other hand, a lower loss is an indicator of redundant temporal batch, which captures temporal events that have already been learned by the model previously. By following this approach, redundant batches can be dropped after calculating the loss and backward propagation can be avoided, which will reduce the overall training time. This approach needs further investigation and has been left for future work. 


{In our experiments, we have shown that there is a trade-off between the performance scaling of matrix multiplications and the model accuracy with different batch sizes. Another future research direction is the optimization of the batch size for a given temporal graph that co-optimizes accuracy and performance scaling. }

\section{Conclusion}
This paper proposes random walk-based temporal graph learning and online training for link prediction and node classification tasks on temporally evolving graphs. The performance evaluation identifies  FNN training as the most expensive operation of the proposed pipeline. The low-dimensional matrix kernels are identified as the main performance bottlenecks during training. Four state-of-the-art matrix kernel parallelization techniques are implemented and evaluated on a large core count machine. The evaluation shows that FNN pipeline acceleration significantly improves the overall performance of the run-time graph learning pipeline.

\section{Acknowledgments}
\label{sec:ack}
This work was funded by the U.S. Government under a grant by the Naval Research Laboratory. This research was also supported by the National Science Foundation under Grant No. CNS-1718481, and in part by the Semiconductor Research Corporation (SRC).

\bibliographystyle{IEEEtran}
\bibliography{references}

\begin{thebibliography}{10}
\providecommand{\url}[1]{#1}
\csname url@samestyle\endcsname
\providecommand{\newblock}{\relax}
\providecommand{\bibinfo}[2]{#2}
\providecommand{\BIBentrySTDinterwordspacing}{\spaceskip=0pt\relax}
\providecommand{\BIBentryALTinterwordstretchfactor}{4}
\providecommand{\BIBentryALTinterwordspacing}{\spaceskip=\fontdimen2\font plus
\BIBentryALTinterwordstretchfactor\fontdimen3\font minus
  \fontdimen4\font\relax}
\providecommand{\BIBforeignlanguage}[2]{{%
\expandafter\ifx\csname l@#1\endcsname\relax
\typeout{** WARNING: IEEEtran.bst: No hyphenation pattern has been}%
\typeout{** loaded for the language `#1'. Using the pattern for}%
\typeout{** the default language instead.}%
\else
\language=\csname l@#1\endcsname
\fi
#2}}
\providecommand{\BIBdecl}{\relax}
\BIBdecl

\bibitem{Scarselli2009TheGN}
F.~Scarselli, M.~Gori, A.~C. Tsoi, M.~Hagenbuchner, and G.~Monfardini, ``The
  graph neural network model,'' \emph{IEEE Transactions on Neural Networks},
  vol.~20, pp. 61--80, 2009.

\bibitem{Hamilton2017InductiveRL}
W.~L. Hamilton, Z.~Ying, and J.~Leskovec, ``Inductive representation learning
  on large graphs,'' in \emph{NIPS}, 2017.

\bibitem{Velickovic2018GraphAN}
P.~Velickovic, G.~Cucurull, A.~Casanova, A.~Romero, P.~Lio’, and Y.~Bengio,
  ``Graph attention networks,'' \emph{ArXiv}, vol. abs/1710.10903, 2018.

\bibitem{Perozzi2014DeepWalkOL}
B.~Perozzi, R.~Al-Rfou, and S.~Skiena, ``Deepwalk: online learning of social
  representations,'' \emph{Proceedings of the 20th ACM SIGKDD international
  conference on Knowledge discovery and data mining}, 2014.

\bibitem{Ahmed2019role2vecRN}
N.~Ahmed, R.~A. Rossi, J.~B. Lee, T.~L. Willke, R.~Zhou, and H.~Eldardiry,
  ``role2vec: Role-based network embeddings,'' 2019.

\bibitem{Ribeiro2017struc2vecLN}
L.~F.~R. Ribeiro, P.~H.~P. Saverese, and D.~R. Figueiredo, ``struc2vec:
  Learning node representations from structural identity,'' \emph{Proceedings
  of the 23rd ACM SIGKDD International Conference on Knowledge Discovery and
  Data Mining}, 2017.

\bibitem{Backstrom2011SupervisedRW}
L.~Backstrom and J.~Leskovec, ``Supervised random walks: predicting and
  recommending links in social networks,'' in \emph{WSDM '11}, 2011.

\bibitem{Bian2020RumorDO}
T.~Bian, X.~Xiao, T.~Xu, P.~Zhao, W.~Huang, Y.~Rong, and J.~Huang, ``Rumor
  detection on social media with bi-directional graph convolutional networks,''
  in \emph{AAAI}, 2020.

\bibitem{Choma2018GraphNN}
N.~Choma, F.~Monti, L.~M. Gerhardt, T.~Palczewski, Z.~Ronaghi, Prabhat,
  W.~Bhimji, M.~M. Bronstein, S.~R. Klein, and J.~Bruna, ``Graph neural
  networks for icecube signal classification,'' \emph{2018 17th IEEE
  International Conference on Machine Learning and Applications (ICMLA)}, pp.
  386--391, 2018.

\bibitem{Duvenaud2015ConvolutionalNO}
D.~K. Duvenaud, D.~Maclaurin, J.~Aguilera-Iparraguirre,
  R.~G{\'o}mez-Bombarelli, T.~D. Hirzel, A.~Aspuru-Guzik, and R.~P. Adams,
  ``Convolutional networks on graphs for learning molecular fingerprints,''
  \emph{ArXiv}, vol. abs/1509.09292, 2015.

\bibitem{Stokes2020ADL}
J.~M. Stokes, K.~Yang, K.~Swanson, W.~Jin, A.~Cubillos-Ruiz, N.~M. Donghia,
  C.~R. MacNair, S.~French, L.~A. Carfrae, Z.~Bloom-Ackermann, V.~M. Tran,
  A.~Chiappino-Pepe, A.~H. Badran, I.~W. Andrews, E.~J. Chory, G.~M. Church,
  E.~D. Brown, T.~Jaakkola, R.~Barzilay, and J.~J. Collins, ``A deep learning
  approach to antibiotic discovery,'' \emph{Cell}, vol. 180, pp. 688--702.e13,
  2020.

\bibitem{Yan2020CharacterizingAU}
M.~Yan, Z.~Chen, L.~Deng, X.~Ye, Z.~Zhang, D.~Fan, and Y.~Xie, ``Characterizing
  and understanding gcns on gpu,'' \emph{IEEE Computer Architecture Letters},
  vol.~19, pp. 22--25, 2020.

\bibitem{Baruah2021GNNMarkAB}
T.~Baruah, K.~Shivdikar, S.~Dong, Y.~Sun, S.~A. Mojumder, K.~Jung, J.~L.
  Abell{\'a}n, Y.~Ukidave, A.~M. Joshi, J.~Kim, and D.~R. Kaeli, ``Gnnmark: A
  benchmark suite to characterize graph neural network training on gpus,''
  \emph{2021 IEEE International Symposium on Performance Analysis of Systems
  and Software (ISPASS)}, pp. 13--23, 2021.

\bibitem{Zhang2020ArchitecturalIO}
Z.~Zhang, J.~Leng, L.~Ma, Y.~Miao, C.~Li, and M.~Guo, ``Architectural
  implications of graph neural networks,'' \emph{IEEE Computer Architecture
  Letters}, vol.~19, pp. 59--62, 2020.

\bibitem{Kipf2017SemiSupervisedCW}
T.~Kipf and M.~Welling, ``Semi-supervised classification with graph
  convolutional networks,'' \emph{ArXiv}, vol. abs/1609.02907, 2017.

\bibitem{Xu2019HowPA}
K.~Xu, W.~Hu, J.~Leskovec, and S.~Jegelka, ``How powerful are graph neural
  networks?'' \emph{ArXiv}, vol. abs/1810.00826, 2019.

\bibitem{Talati2021ADD}
N.~Talati, D.~Jin, H.~Ye, A.~Brahmakshatriya, G.~S. Dasika, S.~P. Amarasinghe,
  T.~N. Mudge, D.~Koutra, and R.~G. Dreslinski, ``A deep dive into
  understanding the random walk-based temporal graph learning,'' \emph{2021
  IEEE International Symposium on Workload Characterization (IISWC)}, pp.
  87--100, 2021.

\bibitem{Nguyen2018ContinuousTimeDN}
G.~H. Nguyen, J.~B. Lee, R.~A. Rossi, N.~Ahmed, E.~Koh, and S.~Kim,
  ``Continuous-time dynamic network embeddings,'' \emph{Companion Proceedings
  of the The Web Conference 2018}, 2018.

\bibitem{mikolov2013efficient}
T.~Mikolov, K.~Chen, G.~Corrado, and J.~Dean, ``Efficient estimation of word
  representations in vector space,'' \emph{arXiv preprint arXiv:1301.3781},
  2013.

\bibitem{Sahoo2018OnlineDL}
D.~Sahoo, Q.~Pham, J.~Lu, and S.~C.~H. Hoi, ``Online deep learning: Learning
  deep neural networks on the fly,'' in \emph{IJCAI}, 2018.

\bibitem{Yoon2018LifelongLW}
J.~Yoon, E.~Yang, J.~Lee, and S.~J. Hwang, ``Lifelong learning with dynamically
  expandable networks,'' \emph{ArXiv}, vol. abs/1708.01547, 2018.

\bibitem{Lee2016DualMemoryDL}
S.-W. Lee, C.~yeon Lee, D.~Kwak, J.~Kim, J.~Kim, and B.-T. Zhang, ``Dual-memory
  deep learning architectures for lifelong learning of everyday human
  behaviors,'' in \emph{IJCAI}, 2016.

\bibitem{Gurevin2021AnEA}
D.~Gurevin, C.~J. Michael, and O.~Khan, ``An efficient algorithm for the
  construction of dynamically updating trajectory networks,'' \emph{2021 IEEE
  High Performance Extreme Computing Conference (HPEC)}, pp. 1--7, 2021.

\bibitem{Heidari2020EvolvingNR}
F.~Heidari and M.~Papagelis, ``Evolving network representation learning based
  on random walks,'' \emph{Applied Network Science}, vol.~5, 2020.

\bibitem{Hamilton2017RepresentationLO}
W.~L. Hamilton, R.~Ying, and J.~Leskovec, ``Representation learning on graphs:
  Methods and applications,'' \emph{ArXiv}, vol. abs/1709.05584, 2017.

\bibitem{Grover2016node2vecSF}
A.~Grover and J.~Leskovec, ``node2vec: Scalable feature learning for
  networks,'' \emph{Proceedings of the 22nd ACM SIGKDD International Conference
  on Knowledge Discovery and Data Mining}, 2016.

\bibitem{LibenNowell2007TheLP}
D.~Liben-Nowell and J.~M. Kleinberg, ``The link-prediction problem for social
  networks,'' \emph{J. Assoc. Inf. Sci. Technol.}, vol.~58, pp. 1019--1031,
  2007.

\bibitem{Sankar2020DySATDN}
A.~Sankar, Y.~Wu, L.~Gou, W.~Zhang, and H.~Yang, ``Dysat: Deep neural
  representation learning on dynamic graphs via self-attention networks,''
  \emph{Proceedings of the 13th International Conference on Web Search and Data
  Mining}, 2020.

\bibitem{Goyal2020dyngraph2vecCN}
P.~Goyal, S.~R. Chhetri, and A.~Canedo, ``dyngraph2vec: Capturing network
  dynamics using dynamic graph representation learning,'' \emph{ArXiv}, vol.
  abs/1809.02657, 2020.

\bibitem{Pareja2020EvolveGCNEG}
A.~Pareja, G.~Domeniconi, J.~Chen, T.~Ma, T.~Suzumura, H.~Kanezashi, T.~Kaler,
  and C.~E. Leisersen, ``Evolvegcn: Evolving graph convolutional networks for
  dynamic graphs,'' in \emph{AAAI}, 2020.

\bibitem{Dunlavy2011TemporalLP}
D.~M. Dunlavy, T.~G. Kolda, and E.~Acar, ``Temporal link prediction using
  matrix and tensor factorizations,'' \emph{ArXiv}, vol. abs/1005.4006, 2011.

\bibitem{Aggarwal2010OnCG}
C.~C. Aggarwal, Y.~Zhao, and P.~S. Yu, ``On clustering graph streams,'' in
  \emph{SDM}, 2010.

\bibitem{Mahmood2018SpatiotemporalAM}
A.~R. Mahmood, S.~Punni, and W.~G. Aref, ``Spatio-temporal access methods: a
  survey (2010 - 2017),'' \emph{GeoInformatica}, vol.~23, pp. 1--36, 2018.

\bibitem{Bulu2008OnTR}
A.~Buluç and J.~R. Gilbert, ``On the representation and multiplication of
  hypersparse matrices,'' \emph{2008 IEEE International Symposium on Parallel
  and Distributed Processing}, pp. 1--11, 2008.

\bibitem{Geijn1997SUMMASU}
R.~A. van~de Geijn and J.~Watts, ``Summa: scalable universal matrix
  multiplication algorithm,'' \emph{Concurr. Pract. Exp.}, vol.~9, pp.
  255--274, 1997.

\bibitem{Gustavson1978TwoFA}
F.~G. Gustavson, ``Two fast algorithms for sparse matrices: Multiplication and
  permuted transposition,'' \emph{ACM Trans. Math. Softw.}, vol.~4, pp.
  250--269, 1978.

\bibitem{Chen2018FastGCNFL}
J.~Chen, T.~Ma, and C.~Xiao, ``Fastgcn: Fast learning with graph convolutional
  networks via importance sampling,'' \emph{ArXiv}, vol. abs/1801.10247, 2018.

\bibitem{Chen2021AUL}
T.~Chen, Y.~Sui, X.~Chen, A.~Zhang, and Z.~Wang, ``A unified lottery ticket
  hypothesis for graph neural networks,'' in \emph{ICML}, 2021.

\bibitem{You2022EarlyBirdGG}
H.~You, Z.~Lu, Z.~Zhou, Y.~Fu, and Y.~Lin, ``Early-bird gcns: Graph-network
  co-optimization towards more efficient gcn training and inference via drawing
  early-bird lottery tickets,'' in \emph{AAAI}, 2022.

\bibitem{Zhang2018ASD}
T.~Zhang, S.~Ye, K.~Zhang, J.~Tang, W.~Wen, M.~Fardad, and Y.~Wang, ``A
  systematic dnn weight pruning framework using alternating direction method of
  multipliers,'' in \emph{ECCV}, 2018.

\bibitem{gurevin2020enabling}
D.~Gurevin, S.~Zhou, L.~Pepin, B.~Li, M.~Bragin, C.~Ding, and F.~Miao,
  ``Enabling retrain-free deep neural network pruning using surrogate
  lagrangian relaxation,'' \emph{arXiv preprint arXiv:2012.10079}, 2020.

\bibitem{Leskovec2014SNAPD}
J.~Leskovec and A.~Krevl, ``\{SNAP Datasets\}: \{Stanford\} large network
  dataset collection,'' 2014.

\bibitem{Paranjape2017MotifsIT}
A.~Paranjape, A.~R. Benson, and J.~Leskovec, ``Motifs in temporal networks,''
  \emph{Proceedings of the Tenth ACM International Conference on Web Search and
  Data Mining}, 2017.

\bibitem{Cunningham2019CreatorGI}
S.~Cunningham and D.~Craig, ``Creator governance in social media
  entertainment,'' \emph{Social Media + Society}, vol.~5, 2019.

\bibitem{Rossi2015TheND}
R.~A. Rossi and N.~Ahmed, ``The network data repository with interactive graph
  analytics and visualization,'' in \emph{AAAI}, 2015.

\bibitem{Shetty2004TheEE}
J.~Shetty and J.~Adibi, ``The enron email dataset database schema and brief
  statistical report,'' 2004.

\bibitem{Xu2019SpatioTemporalAR}
D.~Xu, W.~Cheng, D.~Luo, X.~Liu, and X.~Zhang, ``Spatio-temporal attentive rnn
  for node classification in temporal attributed graphs,'' in \emph{IJCAI},
  2019.

\bibitem{Preti2017TheDF}
M.~G. Preti, T.~A.~W. Bolton, and D.~V.~D. Ville, ``The dynamic functional
  connectome: State-of-the-art and perspectives,'' \emph{NeuroImage}, vol. 160,
  pp. 41--54, 2017.

\end{thebibliography}

\end{document}